\documentclass{article}

\usepackage{microtype}
\usepackage{graphicx}
\usepackage{subfigure}
\usepackage{booktabs} 
\usepackage{hyperref}
\usepackage{amsfonts}
\usepackage{comment}
\usepackage{amsmath}
\usepackage{verbatim}
\usepackage{soul}

\def\ting#1{\textcolor{blue}{Ting: #1}} 

\usepackage[accepted]{icml2020}

\begin{document}

\twocolumn[
\icmltitle{Sharing Models or Coresets: A Study based on Membership Inference Attack
}

\icmlsetsymbol{equal}{*}

\begin{icmlauthorlist}
\icmlauthor{Hanlin Lu}{psu}
\icmlauthor{Changchang Liu}{ibm}
\icmlauthor{Ting He}{equal,psu}
\icmlauthor{Shiqiang Wang}{equal,ibm}
\icmlauthor{Kevin S. Chan}{arl}
\end{icmlauthorlist}

\icmlaffiliation{psu}{Pennsylvania State University, University Park, PA, USA}
\icmlaffiliation{ibm}{IBM T. J. Watson Research Center, Yorktown Heights, NY, USA}
\icmlaffiliation{arl}{Army Research Laboratory, Adelphi, MD, USA}

\icmlcorrespondingauthor{Hanlin Lu}{hzl263@psu.edu}
\icmlcorrespondingauthor{Ting He}{tzh58@psu.edu}

\icmlkeywords{Machine Learning, Member Inference Attack, Coreset, Federated Learning}

\vskip 0.3in
]

\printAffiliationsAndNotice{\icmlEqualContribution} 

\begin{abstract}
Distributed machine learning generally aims at training a global model based on distributed data without collecting all the data to a centralized location, where two different approaches have been proposed: collecting and aggregating local models (federated learning) and collecting and training over representative data summaries (coreset). 
While each approach preserves data privacy to some extent thanks to not sharing the raw data, the exact extent of protection is unclear under sophisticated attacks that try to infer the raw data from the shared information. 
We present the first comparison between the two approaches in terms of target model accuracy, communication cost, and data privacy, where the last is measured by the accuracy of a state-of-the-art attack strategy called the membership inference attack. 
Our experiments quantify the accuracy-privacy-cost tradeoff of each approach, and reveal a nontrivial comparison that can be used to guide the design of model training processes. 
\end{abstract}

\section{Introduction}

Machine learning (ML) has been the force of innovation behind many sensor-driven intelligent systems such as autonomous vehicles, personalized healthcare,  precision agriculture, machine health monitoring, environmental tracking, and infrastructure security. In many applications, because of the limited observability at a single sensor (e.g., due to limited sensing range or angle), machine learning needs to be performed on the data from multiple data sources.  However, due to limitations on network bandwidth, power consumption, and data privacy, data sources often cannot upload their raw data. Instead, they share the information in their local data using various distributed learning approaches, where two of the most well-known approaches are: (i) sharing models trained on the local data (e.g., federated learning~\cite{konecny16NIPS}), or (ii) sharing representative summaries of the local data (e.g., coreset~\cite{Feldman11NIPS}). 

While each of these approaches has been studied separately, to our knowledge, a comprehensive comparison between them is still missing. In this work, we take a first step towards filling this gap by comparing the federated learning approach and the coreset-based approach in terms of (1) the \emph{accuracy} of the target machine learning model we want to train, (2) the \emph{communication cost} during training, and (3) the \emph{leakage} of the private training data. In particular, although neither approach will require the data sources to directly share their data, it has been shown in \cite{shokri2017membership} that models derived from a dataset can be used to infer the membership of the dataset (i.e., whether or not a given data record is contained in the dataset), known as the \emph{membership inference attack (MIA)}. Since the coreset can also be viewed as a model, we can thus use the accuracy of MIA as a practical measure of the privacy leakage.

\subsection{Related Work}

Federated learning~\cite{konecny16NIPS} and coreset~\cite{Feldman11NIPS} can both be used to train ML models over distributed data without collecting the raw data to a central location. Each approach has been independently studied, such as adaptive federated learning~\cite{wang2019adaptive}, 
communication-efficient coreset construction~\cite{Balcan13NIPS}, and the robustness of coreset in supporting multiple ML models~\cite{lu2019robust}. However, a comprehensive comparison between them is still missing. In this work, we perform an initial comparison with focus on data privacy. 

{ML models, which are derived from the training data, inevitably contain information of the records in the training dataset. By utilizing this characteristic, various attacks have been proposed to infer private information of the training data from the existing ML models including neural networks. These attacks can be generally classified into two categories: \emph{model inversion attacks} and \emph{membership inference attacks.} } 

{Model inversion attacks try to recover the training data by observing the model predictions of the target model. The first model inversion attack was proposed by Fredrikson et al. \cite{fredrikson2015model} that exploits a given face label to reconstruct the face of an individual belonging to the training data of a face recognition system, which has been generalized to attack the deep neural network models \cite{zhang2019secret}, generative models \cite{hayes2019logan}, and federated learning \cite{geiping2020inverting}. However, the model inversion attacks are usually limited to certain applications which may not be applicable to tasks other than face recognition \cite{rahman2018membership}.} 

{In membership inference attack (MIA) \cite{shokri2017membership,salem2018ml,nasr2019comprehensive,sablayrolles2019white}, the adversary tries to infer whether a given input data record belongs to the original training data of the target model or not. This attack, proposed in \cite{shokri2017membership}, can be applied to any type of ML models and even under the black-box setting, where only the model output (but not the internal model parameters) is known to the adversary. Salem et al. \cite{salem2018ml} relaxed the assumptions for shadow models to broaden the applicability of MIA in practice. Sablayrolles et al. \cite{sablayrolles2019white} proposed the \emph{Bayes} optimal strategies for MIA, by assuming certain distributions of model parameters. Nasr et al. \cite{nasr2019comprehensive} has further extended MIA to federated learning in order to quantify the privacy leakage in the distributed setting. }\looseness=-1

\subsection{Summary of Contribution}

We aim to compare the approach of federated learning with the approach of sharing coresets, with focus on the privacy risks imposed by MIA. {Specifically:} 
\begin{itemize}
    \item {We are the first to systematically compare the approaches of federated learning and coreset-based learning in accuracy, privacy, and communication cost.}
    \item {We propose the first MIA against distributed coreset.}
    \item {Through experiments on real datasets, we not only quantify the accuracy-privacy-cost tradeoff of each approach, but also reveal that which approach achieves a better tradeoff can be different under different accuracy requirements.} 
\end{itemize}

\textbf{Roadmap.} Section~\ref{sec:Overview of MIA} reviews the high-level idea of MIA. Section~\ref{sec:MIA on Federated Learning and Coreset} defines it in detail for federated learning and coreset. Sections~\ref{sec:Experiment Design} and \ref{sec:Evaluation} present our experiment design and results. Then Section~\ref{sec:Conclusion} concludes the paper with discussions. \looseness=-1

\section{Overview of Membership Inference Attack}\label{sec:Overview of MIA}

We start by reviewing the basic idea of MIA as proposed in \cite{shokri2017membership}, based on the following models:\looseness=-1

\emph{-- Target model:} Target model is the machine learning model trained on private data and then revealed to the public (including the adversary). Given a data record $\boldsymbol{d}$, let $f_{target}(\boldsymbol{d})$ denote the output of the target model, where $f_{target}(\cdot)$ is referred to as the prediction function. For example, if the target model is a neural net trained to predict the label in a set of all possible labels $\boldsymbol{D}_{target}$, then $f_{target}(\boldsymbol{d})$ will be a $|\boldsymbol{D}_{target}|$-dimensional vector denoting the predicted probabilities for $\boldsymbol{d}$ to have each of the labels in $\boldsymbol{D}_{target}$. Let $\boldsymbol{D}_{target}^{train}$ and $\boldsymbol{D}_{target}^{test}$ denote the training/testing dataset of the target model. 

\emph{-- Shadow model:} Shadow model is a machine learning model of the same type as the target model, constructed by the adversary to mirror the target model. Let $\boldsymbol{D}^{train}_{shadow}$ and $\boldsymbol{D}^{test}_{shadow}$ denote the training/testing data of the shadow model, and $f_{shadow}(\cdot)$ denote the prediction function. 

\emph{-- Attack model:} Attack model is a binary classifier designed by the adversary to infer the membership of data records of interest with respect to the training dataset of the target model, which will have different structures according to different target models. 

\emph{Workflow:} {The key intuition of MIA is to exploit the difference in how the target model responds to data records that are in its training data versus data records that are not. At a high level, MIA  works in three steps: (1) First, the adversary constructs shadow models to mirror the behavior of the target model. (2) Then, the shadow models are used to generate training data for the attack model. Specifically, the prediction vector on each training data record of the shadow model, i.e., $f_{shadow}(\boldsymbol{d})$ for $\boldsymbol{d} \in \boldsymbol{D}^{train}_{shadow}$, is labeled as \emph{``IN''} (in the training data), and the prediction vector on each testing data record, i.e., $f_{shadow}(\boldsymbol{d})$ for $\boldsymbol{d}\in \boldsymbol{D}^{test}_{shadow}$, is labeled as \emph{``OUT''} (not in the training data). (3) Finally, these prediction vectors together with their associated ``IN/OUT'' labels are used to train the attack model, which can then be used to infer whether a given data record is in the training data of the target model.}

\emph{Enhancement:} 
The accuracy of MIA depends on how well the shadow model approximates the target model and how well $\boldsymbol{D}^{train}_{shadow}$ approximates $\boldsymbol{D}^{train}_{target}$. While several methods of generating $\boldsymbol{D}^{train}_{shadow}$ have been proposed in~\cite{shokri2017membership}, how well each method approximates $\boldsymbol{D}^{train}_{target}$ can vary substantially from case to case. To obtain a conservative estimation of the privacy leakage, we assume a stronger adversary who \emph{has access to part of the target model's training/testing data.} This assumption is the same as the supervised white box setting in \cite{nasr2019comprehensive}. Specifically, let $\boldsymbol{S}^{train}_{target}\subset \boldsymbol{D}^{train}_{target}$ and $\boldsymbol{S}^{test}_{target}\subset \boldsymbol{D}^{test}_{target}$ denote the subsets of the training/testing data leaked to the adversary. Then the adversary will know that each $f_{target}(\boldsymbol{d})$ for $\boldsymbol{d}\in \boldsymbol{S}^{train}_{target}$ has the label ``IN'' (in the training data), and each $f_{target}(\boldsymbol{d})$ for $\boldsymbol{d} \in \boldsymbol{S}^{test}_{target}$ has the label ``OUT'' (not in the training data), which provides the following training dataset for the attack model:
\begin{align}\label{eq:D^train_attack}
    \boldsymbol{D}_{attack}^{train} =\, & \{(f_{target}(\boldsymbol{d}), \mbox{IN})\}_{\boldsymbol{d}\in \boldsymbol{S}_{target}^{train}} \nonumber \\
    &\cup \{f_{target}(\boldsymbol{d}), \mbox{OUT})\}_{\boldsymbol{d}\in \boldsymbol{S}_{target}^{test}}.
\end{align}
Therefore, under this assumption, this stronger adversary no longer needs to construct shadow models, which are directly replaced by the target model. The focus is thus on designing the attack model. To attack federated learning, we will apply the design in \cite{nasr2019comprehensive}. To attack coreset-based learning, we will design our own attack model as explained later in Section~\ref{subsubsec:attack on coreset}. 


\section{Membership Inference Attack under Federated or Coreset-based Learning}\label{sec:MIA on Federated Learning and Coreset}

In this section, we will provide details about the design of MIA under federated learning and coreset-based learning, respectively. For both approaches, we consider a setting with a server interested in learning the target model over data distributed across $N$ nodes. We assume that the adversary can eavesdrop all the communications between the server and the nodes. In the sequel, we will use neural network as the target model, which is known to be particularly vulnerable to MIA~\cite{nasr2019comprehensive}. 

\subsection{Attack on Federated Learning}

\subsubsection{Federated Learning}
Assuming that stochastic gradient descent is used to train the target model, we consider a version of federated learning with one global aggregation per training epoch, i.e., all the nodes will report their local models to a central server after each step of local gradient descent, and the server will aggregate the models by taking a weighted average and send the resulting model to the nodes. This version of federated learning is known to give a target model that is equivalent to the model trained by gradient descent on the union of all the local training data~\cite{wang2019adaptive}. 

\subsubsection{Attack Model} \label{sec: attack model against fed learning}

\begin{figure}[ht]
\vskip 0in
\begin{center}
\centerline{\includegraphics[width=0.7\columnwidth]{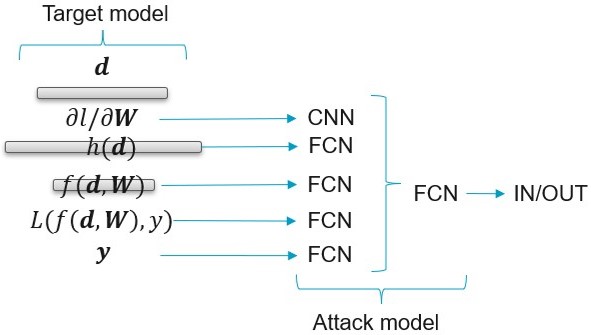}}
\vspace{-1em}
\caption{Attack model against federated learning}
\label{fig: attack_fed_learning}
\end{center}
\vskip -0.2in
\end{figure}

In contrast to the black-box attack in the centralized setting~\cite{shokri2017membership}, MIA under federated learning is a white-box attack, since all the information exchanged during the training process is also received by the adversary. That is, the adversary observes not only the outputs of the target model, but also the internal parameters of the target model and all the intermediate results during training. To design the attack in the white-box setting, we apply the methods from \cite{nasr2019comprehensive}. Let $f(\boldsymbol{d}, \boldsymbol{W})$ and $L(f(\boldsymbol{d}, \boldsymbol{W}), y)$ denote the prediction function and the loss function associated with the target model, where $\boldsymbol{d}$ is an input data record, $y$ is the corresponding label, and $\boldsymbol{W}$ is the vector of model parameters. Then the attack model will have 5 components of input from each training epoch ($n$: number of layers of the target model):
\begin{itemize}
    \item The gradients of the loss function $\{\partial L / \partial \boldsymbol{W}_i \}_{i=1}^n$ from all the layers of the target model ($\boldsymbol{W}_i$: parameters at layer $i$); 
    \item The activation vectors $\{h_i(\boldsymbol{d})\}_{i=1}^{n-1}$ from all hidden layers ($h_i(\boldsymbol{d})$: output of layer $i$ for input record $\boldsymbol{d}$);
    \item The output $f(\boldsymbol{d}, \boldsymbol{W})$ of the target model;
    \item The one-hot encoding ($\boldsymbol{y}$) of label $y$\footnote{One-hot encoding $\boldsymbol{y}$ of a label $i\in \{1,\ldots,N\}$ is a length-$N$ binary vector with the $i$-th entry being `1' and all the rest being `0' \cite{one-hot}.};
    \item The loss of the target model $L(f(\boldsymbol{d}, \boldsymbol{W}), y)$; 
\end{itemize}
The attack model is a hierarchy of neural networks as illustrated in Figure~\ref{fig: attack_fed_learning}. First, each component is fed into a fully connected neural network (FCN),  except for the gradient component which is fed into a convolutional neural network (CNN). Then, the outputs of all the 5 neural networks are concatenated and passed to an FCN with IN/OUT probabilities as the final output. See \cite{nasr2019comprehensive} for more details.\looseness=-1 

\subsection{Attack on Coreset-based Learning}
\subsubsection{Coreset Construction}
To construct a coreset over distributed data, we apply the \emph{distributed robust coreset construction} (\textit{DRCC}) algorithm from \cite{lu2019robust}, which is proved to support multiple machine learning models with guaranteed performance. The \textit{DRCC} algorithm works in three steps:
\begin{enumerate}
    \item Each node calculates its own $k$-means centers for several $k$ values and 
    returns the $k$-means costs to the central server;
    \item Based on the local $k$-means costs, the central server computes the number of local $k$-means centers and the number of local samples for each node according to a desired total coreset size, and communicates these to the nodes;
    \item After receiving the parameters from the server, each node will calculate and report its local coreset, which consists of two parts: (1) the local $k$-means centers, (2) the local samples randomly drawn from the local dataset with probabilities proportional to the squared Euclidean distance from each data record to the nearest local $k$-means center. 
\end{enumerate}

\subsubsection{Attack model} \label{subsubsec:attack on coreset}

From the coreset construction algorithm, clearly the main information exposed to the adversary is the coreset itself, which is a union of the local coresets. 
Since the local samples are from the training data and thus expose part of the training dataset, we will include them in $S^{train}_{target}$\footnotemark. 
%
The focus is on designing an attack model based on the local $k$-means centers to infer the membership of the unexposed training data. To our knowledge, how to design the attack model for MIA on $k$-means has not been studied before. \footnotetext{We also consider constructing coreset with $k$-means centers only, whose results are given in Appendix \ref{appendix: results without local samples}. }

According to the design principle of MIA (Section~\ref{sec:Overview of MIA}), the first step is to define a proper prediction function $f_{target}(\cdot)$ for $k$-means. Although the natural output of $k$-means for an input data record $\boldsymbol{d}$ should be the index of the nearest center (i.e., the cluster index of $\boldsymbol{d}$), this information is too little for membership inference.  
We thus propose to define $f_{target}$ to include the distances from the input data record to all the $k$-means centers: $f_{target}(\boldsymbol{d}) = (\|\boldsymbol{d}-\boldsymbol{c}_1\|, \ldots, \|\boldsymbol{d}-\boldsymbol{c}_{\hat{k}}\|)$, where $\|\cdot\|$ denotes the Euclidean distance, and $\{\boldsymbol{c}_i\}_{i=1}^{\hat{k}}$ is the set of local centers reported from all the nodes. 



\begin{figure}
\centering
\begin{minipage}{.24\textwidth}
\centerline{
  \includegraphics[width=0.8\textwidth]{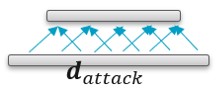}}
  \centerline{(a) concatenation }
\end{minipage}\hfill
\begin{minipage}{.24\textwidth}
\centerline{
  \includegraphics[width=0.8\textwidth]{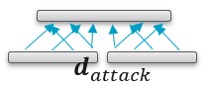}}
  \centerline{(b) hierarchical }
\end{minipage}
\vspace{-1em}
\caption{Two architectures of attack model for coreset ($N=2$)}  
\label{fig: two input}
 \vspace{-1em}
\end{figure}

We then use such distance vectors together with their IN/OUT labels generated as in Equation~\eqref{eq:D^train_attack} to train a neural network as the attack model. As the features (i.e., distances to local centers) are naturally partitioned into $N$ sets corresponding to the $N$ nodes,  we  propose two architectures for the attack model: (1) \textit{concatenation} (Figure \ref{fig: two input} (a)), which is to concatenate all the features and pass them into an FCN; (2) \textit{hierarchical} (Figure \ref{fig: two input} (b)), which is to pass features from different nodes into different neurons of the second layer, whose outputs are then concatenated and passed into higher layers which form an FCN. The intuition behind the hierarchical design is that by feeding features from each node to only ($1/N$)-th of the neurons in the second layer, we can reduce the number of parameters to learn and thus potentially train the attack model better on limited data. \looseness=-1

\section{Experiment Design}\label{sec:Experiment Design}

\subsection{Datasets}

We used two datasets in our experiments, both of which have been used for MIA 
\cite{shokri2017membership}: \textit{Locations} dataset, and \textit{Purchase100} dataset.

\emph{Locations}: Locations dataset was created by collecting Foursquare social network's location ``check-ins'' in Bangkok from April 2012 to September 2013. It contains 5010 records with 446 binary features each, representing whether a user visited certain places. The dataset has been clustered into 30 clusters, and the cluster indices represent 30 different geosocial types. The target model for this dataset aims at predicting the geosocial type of a new user.  

\emph{Purchase100}: Purchase100 dataset was simplified from the Kaggle’s ``acquire valued shoppers'' challenge dataset. It contains 197,324 users' purchase history in one year, with 600 binary features each denoting whether or not a user purchased a certain product. The dataset has been clustered into 100 clusters, where the cluster indices represent different purchasing styles. The target model for this dataset aims at predicting the purchasing style of a new user. 

\subsection{Experiment Setup}

All the experiments are based on a distributed setting, in which there are two nodes each holding half of the training data. For Locations dataset, we randomly select 3,000 data records as the target model's training data and leave the rest as testing data. 
For Purchase100 dataset, we randomly select 30,000 data records as the target model's training data, 
and another 20,000 data records as the target model's testing data. 
%
As explained in Section~\ref{sec:Overview of MIA}, we assume that part of the target model's training/testing data has been exposed to the adversary for a conservative evaluation of the privacy leakage. Specifically, let $\boldsymbol{S}_{target}^{train}\subset \boldsymbol{D}_{target}^{train}$ and $\boldsymbol{S}_{target}^{test}\subset \boldsymbol{D}_{target}^{test}$ denote the subsets of the target model's training and testing data that are exposed to the adversary. The goal of MIA is to use this information together with the information exposed during the target model training process (e.g., gradients, coreset) to infer the membership of data records in the unexposed part of the training data (i.e., $\boldsymbol{D}_{target}^{train}\setminus \boldsymbol{S}_{target}^{train}$). In our experiments, we use ${\boldsymbol{S}'}_{target}^{train}\subset \boldsymbol{D}_{target}^{train}$ as ``IN'' samples and ${\boldsymbol{S}'}_{target}^{test}\subset \boldsymbol{D}_{target}^{test}$ as ``OUT'' samples to test the accuracy of the attack model, satisfying 
\begin{align}
    \boldsymbol{S}_{target}^{train} \cap {\boldsymbol{S}'}_{target}^{train} = \emptyset,~~~\boldsymbol{S}_{target}^{test} \cap {\boldsymbol{S}'}_{target}^{test} = \emptyset.
\end{align}
For Locations, we set $|\boldsymbol{S}_{target}^{train}|=1,750$, $|\boldsymbol{S}_{target}^{test}|=1750$, $|{\boldsymbol{S}'}_{target}^{train}|=250$, and $|{\boldsymbol{S}'}_{target}^{test}|=250$. For Purchase100, we set $|\boldsymbol{S}_{target}^{train}|=6,000$, $|\boldsymbol{S}_{target}^{test}|=6,000$, $|{\boldsymbol{S}'}_{target}^{train}|=2,000$, and $|{\boldsymbol{S}'}_{target}^{test}|=2,000$.

\emph{Evaluation metrics:} In order to compare the different approaches of training the target model (federated learning vs. coreset), we propose the following evaluation metrics:
\begin{itemize}
    \item the accuracy of the target model (`accuracy'), measured by the target model's testing accuracy,
    \item the privacy leakage (`leakage'), measured by the attack model's testing accuracy, and
    \item the communication cost (`cost'), measured by number of scalars transmitted by all the nodes during training. \looseness=-1 
\end{itemize}
As both approaches have design parameters to control the tradeoff between these metrics (\#training epochs for the federated learning approach and coreset size for the coreset-based approach), we will first tune the design parameters to align the quality of the target models trained by each approach and then compare the other metrics.

\section{Experiment Results} \label{sec:Evaluation}

\subsection{Results on Locations Dataset}

We design the target model as a 4-layer FCN with 256 and 128 neurons in the first and the second hidden layers, and 30 neurons in the last layer, whose output represents the predicted probabilities for the input data record to belong to each of the 30 geosocial types. When trained on the raw training data, this model yields a testing accuracy of $70\%$.\looseness=-1 

\subsubsection{Parameter Setting} \label{sec: locations parameter setting}

\begin{figure}
\centering
\begin{minipage}{.24\textwidth}
\centerline{
  \includegraphics[width=1\textwidth]{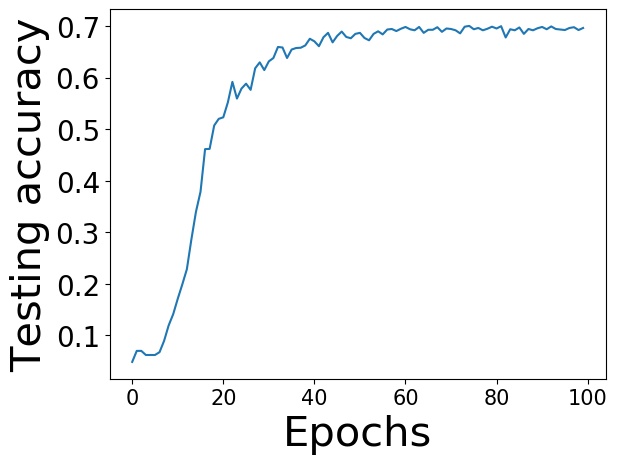}}
  \centerline{(a) vs. \#epochs }
\end{minipage}\hfill
\begin{minipage}{.24\textwidth}
\centerline{
  \includegraphics[width=1\textwidth]{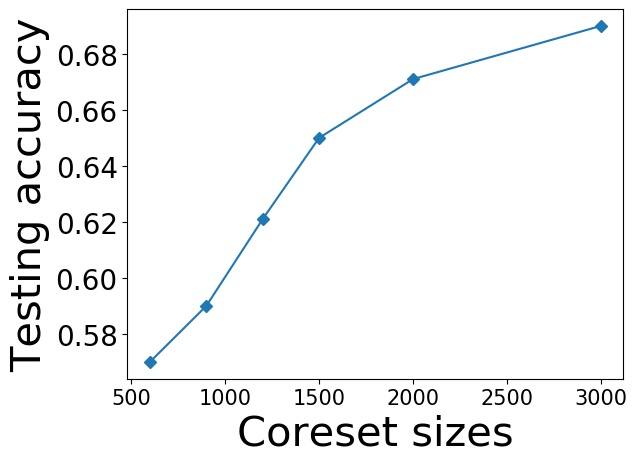}}
  \centerline{(b) vs. coreset size }
\end{minipage}
\vspace{-1em}
\caption{Testing accuracy of the target model vs. \#epochs and coreset size (Locations dataset) }
\label{fig: locations Testing accuracy vs epochs and coresey sizes}
 \vspace{-1em}
\end{figure}

As shown in Figure~\ref{fig: locations Testing accuracy vs epochs and coresey sizes}, both approaches of training the target model on distributed training data exhibit a concave and increasing accuracy as a function of the design parameters. However, the coreset-based approach requires a coreset size almost equal to the size of the entire training dataset to obtain a target model as accurate as the model trained on the raw training data. Therefore, if we want to fully preserve the accuracy of the target model while having some level of protection on the training data, only federated learning is applicable.  Meanwhile, if we can tolerate a little loss of accuracy, then both approaches become applicable. 

\emph{Federated Learning:} 
To test the full accuracy scenario, we run federated learning until convergence, which takes 100 epochs. To test the partial accuracy scenario, we manually terminate the training process at epoch 33, which yields a target model with testing accuracy $65.9\%$ as shown in Figure~\ref{fig: locations Testing accuracy vs epochs and coresey sizes}~(a). 
Our attack model is constructed based on the approach discussed in Section~\ref{sec: attack model against fed learning}, which takes 5 components as inputs: gradients, hidden layer outputs, final outputs, loss and one-hot encoding of labels. Due to the large number of such inputs over all the training epochs, we only use inputs from 5 epochs evenly distributed in the training process to train the attack model. Moreover, for the first two components, we only utilize the last layer's gradients and the last hidden layer's outputs. 
The way we apply MIA and the parameters we use are consistent with those proposed in Appendix~A of \cite{nasr2019comprehensive}. 

\emph{Coreset-based Learning:}
As shown in Figure~\ref{fig: locations Testing accuracy vs epochs and coresey sizes}~(b), the coreset degenerates into the entire training dataset to achieve full accuracy. The focus is thus on the partial accuracy scenario, where we set the coreset size to 1,500, containing $100$ random samples (considered part of the exposed training data $S^{train}_{target}$) and $700$ local centers from each node. This coreset yields a target model with testing accuracy matching that trained by federated learning after 33 epochs. The result for another coreset containing no random sample can be found in the appendix.

\subsubsection{Results}\label{subsubsec:Results - Locations}

The results are shown in Table~\ref{tab:results on Locations}. We see that: (i) in the full accuracy scenario ($70\%$ accuracy), sharing coreset degenerates into sharing the raw training data, and hence the federated learning approach is more desirable in terms of privacy protection; (ii) in the partial accuracy scenario ($\approx 65\%$ accuracy), sharing coreset no longer causes more privacy leakage than federated learning, while having the advantage of incurring a much lower communication cost. 

\begin{table}[tb]
\small
\renewcommand{\arraystretch}{1.3}
\caption{Results on Locations Dataset} \label{tab:results on Locations}
\centering
\begin{tabular}{c||c|c|c}
\hline
approach & accuracy & leakage & cost \\
\hline
\begin{tabular}{c}
federated learning\\
(33 epochs)
\end{tabular} & $65.9\%$ & $57.4\%$  &  19,958,136 \\
\hline
coreset (size 1,500) & $65.0\%$ & $55\%\footnotemark$ &  700,500 \\
\hline
\hline
\begin{tabular}{c}
federated learning\\
(100 epochs)
\end{tabular} & $70\%$ & $73.8\%$ & 60,479,200 \\
\hline 
coreset (size 3,000) & $70\%$ & $100\%$ & 1,401,000\\
\hline
\end{tabular}
 \vspace{-0.2in}
\end{table}
\footnotetext{This is achieved under the concatenation architecture. Under the hierarchical architecture, the attack accuracy is $51.6\%$. }

\subsection{Results on Purchase100 Dataset} 

We design the target model as a 6-layer FCN with 1024, 512, 256, and 128 neurons in the hidden layers, and 100 neurons in the last layer, whose output denotes the predicted probabilities for the input data record to belong to each of the 100 purchasing styles. When trained on the raw training data using stochastic gradient descent, this target model yields a testing accuracy of $83.4\%$. 

\subsubsection{Parameter Setting}\label{subsubsec: Purchast100 parameter setting}

\begin{figure}
\centering
\begin{minipage}{.24\textwidth}
\centerline{
  \includegraphics[width=1\textwidth]{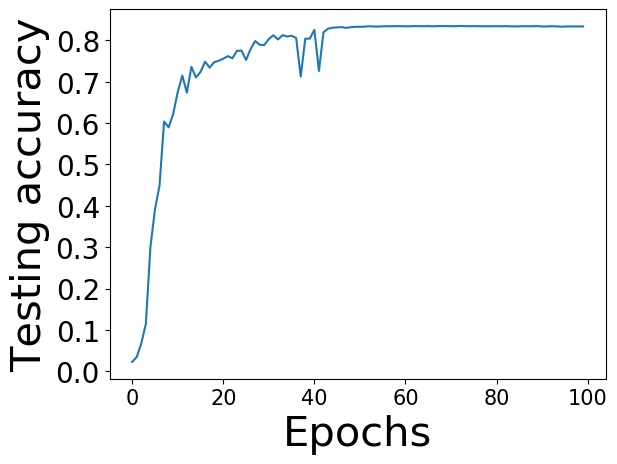}}
  \centerline{(a) vs. \#epochs }
\end{minipage}\hfill
\begin{minipage}{.24\textwidth}
\centerline{
  \includegraphics[width=1\textwidth]{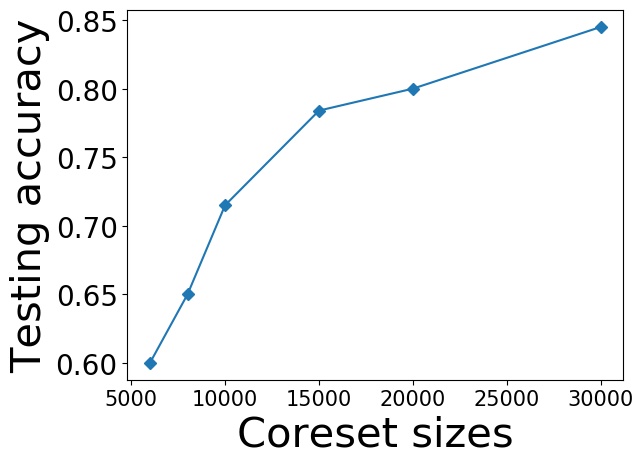}}
  \centerline{(b) vs. coreset size }
\end{minipage}
\vspace{-1em}
\caption{Testing accuracy of the target model vs. \#epochs and coreset size (Purchase100 dataset)}  
\label{fig: purchase100 Testing accuracy vs epochs and coresey sizes}
\end{figure}

Figure~\ref{fig: purchase100 Testing accuracy vs epochs and coresey sizes} shows the testing accuracy of the target model trained by each approach as a function of its design parameter. We observe a similar trend as in Figure~\ref{fig: locations Testing accuracy vs epochs and coresey sizes}, where the coreset-based approach again requires a coreset as large as the entire training dataset to achieve the full accuracy.  

\emph{Federated Learning:} 
To test the full accuracy scenario, we run federated learning for 100 epochs, which guarantees convergence. To test the partial accuracy scenario, we terminate the training process at epoch 30, which gives a target model with $79.8\%$ testing accuracy as shown in Figure~\ref{fig: purchase100 Testing accuracy vs epochs and coresey sizes}~(a). 
The attack model is designed in the same way as that for the Locations dataset. In particular, following the approach in \cite{nasr2019comprehensive}, we train the attack model on the inputs from 5 epochs evenly distributed in the training process, and from each of these epochs, we only use the last layer's gradients and the last hidden layer's outputs, in addition to the final outputs, loss, and one-hot encoding of labels, to train the attack model. 

\emph{Coreset-based Learning:}
As the coreset offers no privacy protection in the full accuracy scenario, we focus on the partial accuracy scenario. According to Figure~\ref{fig: purchase100 Testing accuracy vs epochs and coresey sizes}~(b), we set the coreset size to 15,000, including 5,000 random samples (considered part of $S^{train}_{target}$) and 5,000 local centers from each node. The target model trained on this coreset has a similar testing accuracy as the target model trained by federated learning after 30 epochs. See the appendix for results on another coreset with no random sample.

\subsubsection{Results}\label{subsubsec:Results - Purchase100}

The results, shown in Table~\ref{tab:results on Purchase100}, indicate similar comparisons as in Section~\ref{subsubsec:Results - Locations}, i.e., (i) in the full accuracy scenario ($83.4\%$), the federated learning approach is more desirable as it still provides some privacy protection (although at a much higher communication cost); (ii) if we can tolerate mild loss of accuracy ($\approx 79\%$), then sharing coreset becomes more desirable as it achieves the same accuracy as federated learning with less privacy leakage and a much lower communication cost. 

\begin{table}[tb]
\small
\renewcommand{\arraystretch}{1.3}
\caption{Results on Purchase100 Dataset} \label{tab:results on Purchase100}
\centering
\begin{tabular}{c||c|c|c}
\hline
approach & accuracy & leakage & cost \\
\hline
\begin{tabular}{c}
federated learning\\
(30 epochs)
\end{tabular} & $79.8\%$ & $59.8\%$  &  158,081,760 \\
\hline
coreset (size 15,000) & $78.4\%$ & $51\%\footnotemark$ &  9,015,000 \\
\hline
\hline
\begin{tabular}{c}
federated learning\\
(100 epochs)
\end{tabular} & $83.4\%$ & $72.0\%$ &  526,939,200  \\
\hline 
coreset (size 30,000) & $83.4\%$ & $100\%$ & 18,030,000 \\
\hline
\end{tabular}
 \vspace{-0.2in}
\end{table}
\footnotetext{This is achieved under the hierarchical architecture. Under the concatenation architecture, the attack accuracy is $50\%$. }

\section{Concluding Discussions}\label{sec:Conclusion}

In this work, we present the first systematic comparison between federated learning and coreset-based learning in terms of the target model accuracy, the privacy leakage under MIA, and the communication cost. By experimenting with state-of-the-art attack models and real datasets, we reveal a nontrivial comparison between the two approaches, where federated learning is better at protecting the privacy of the training data when we need a target model of the maximum accuracy, whereas coreset-based learning can provide comparable privacy at a much lower communication cost when we can tolerate a mild loss of target model accuracy. 
Our observation is consistent with the existing result in \cite{nasr2019comprehensive}, that the gradients during the training process can expose the membership of the training data, as federated learning exposes this information but coreset does not. \looseness=-1

We have only compared the original versions of federated learning and coreset-based learning that do not consider privacy. 
Recently, several defenses have been proposed to mitigate MIA on machine learning models, which either insert noise to the target model training process by using differential privacy or adversarial learning \cite{jia2019memguard}, or regulate the training process to reduce overfitting \cite{nasr2018machine}. There have also been defenses proposed to enhance the privacy protection for coresets~\cite{Feldman17IPSN}. 
Therefore, investigating how the privacy-aware versions of these approaches compare in terms of accuracy, privacy, and cost can be an interesting future direction. 

Methods for optimizing federated learning \cite{konecny16NIPS,wang2019adaptive,han2020adaptive} can also be incorporated into this study in the future, which can significantly reduce the communication cost of federated learning.
\looseness=-1

\section*{Acknowledgements}
This research was sponsored by the U.S. Army Research Laboratory and the U.K. Ministry of Defence under Agreement Number W911NF-16-3-0001. The views and conclusions contained in this document are those of the authors and should not be interpreted as representing the official policies, either expressed or implied, of the U.S. Army Research Laboratory, the U.S. Government, the U.K. Ministry of Defence or the U.K. Government. The U.S. and U.K. Governments are authorized to reproduce and distribute reprints for Government purposes notwithstanding any copyright notation hereon.

\bibliography{mybib}
\bibliographystyle{icml2020}

\clearpage

\appendix
\section{Appendix: Additional Evaluations} \label{appendix: results without local samples}
In addition to the typical construction of distributed coreset, we have also evaluated the setting when distributed coreset consists of $k$-means centers only. By constructing coreset in this way, there will be no data record from the original training dataset exposed directly through the coreset. Specifically, the results are given below. 

\subsection{Locations dataset}
All parameters used in this section for Locations dataset are the same as in Section \ref{sec: locations parameter setting} except the structure of distributed coreset. The coreset has $1,500$ $k$-means centers only (750 per node), instead of both $k$-means centers and random samples. The results comparing federated learning and this distributed coreset 
are given in Table \ref{tab:results on Locations with kmeans centers only}, which show a qualitatively similar comparison as in Section~\ref{subsubsec:Results - Locations}.

\subsection{Purchase100 dataset} 
Similarly, we collect $7,500$ $k$-means centers from each node to form a distributed coreset, based on which the target model's testing accuracy is now $75.0\%$. In order to align with this testing accuracy, we manually pause federated learning at epoch 20 and obtain $75.1\%$ testing accuracy. The remaining parameters and designs are the same as in Section~\ref{subsubsec: Purchast100 parameter setting}. The results, shown in Table~\ref{tab:results on Purchase100 with kmeans centers only}, are qualitatively similar to those in Section~\ref{subsubsec:Results - Purchase100}.

\subsection{Discussions}
According to the results as shown in tables above, although the gap in MIA accuracy between the two approaches is smaller, we can still draw similar conclusions as before, while not directly revealing any training data through the coreset. That is, if we can tolerate a moderate degradation of the target model's performance, then we can construct distributed coreset with $k$-means centers only, which will provide better privacy and less communication cost than federated learning. 

\newpage

\begin{table}[tb]
\small
\renewcommand{\arraystretch}{1.3}
\caption{Results on Locations Dataset with $k$-means centers only} \label{tab:results on Locations with kmeans centers only}
\centering
\begin{tabular}{c||c|c|c}
\hline
approach & accuracy & leakage & cost \\
\hline
\begin{tabular}{c}
federated learning\\
(33 epochs)
\end{tabular} & $65.9\%$ & $57.4\%$  &  19,958,136 \\
\hline
\begin{tabular}{c}
     coreset \\
     (size $1,500$,  \\
      with $k$-means \\
      centers only)
\end{tabular}  & $64.48\%$ & $54.4\%\footnotemark$ &  700,500 \\
\hline
\end{tabular}
\end{table}
\footnotetext{This is achieved under the concatenation architecture. Under the hierarchical architecture, the attack accuracy is $52.8\%$. }

\begin{table}[tb]
\small
\renewcommand{\arraystretch}{1.3}
\caption{Results on Purchase100 Dataset with $k$-means centers only} \label{tab:results on Purchase100 with kmeans centers only}
\centering
\begin{tabular}{c||c|c|c}
\hline
approach & accuracy & leakage & cost \\
\hline
\begin{tabular}{c}
federated learning\\
(20 epochs)
\end{tabular} & $75.1\%$ & $55.4\%$  &  105,387,840 \\
\hline
\begin{tabular}{c}
     coreset \\
     (size $15,000$,  \\
      with $k$-means \\
      centers only)
\end{tabular}  & $75.0\%$ & $51.9\%\footnotemark$ &  9,015,000 \\
\hline
\end{tabular}
\end{table}
\footnotetext{This is achieved under the hierarchical architecture. Under the concatenation architecture, the attack accuracy is $51.4\%$. }

\end{document}